\pdfoutput=1

\documentclass[11pt]{article}

\usepackage[preprint]{acl}

\usepackage{times}
\usepackage{latexsym}

\usepackage[T1]{fontenc}

\usepackage[utf8]{inputenc}

\usepackage{microtype}

\usepackage{inconsolata}

\usepackage{graphicx}

\usepackage{hyperref}
\usepackage{url}            
\usepackage{booktabs}       
\usepackage{amsfonts}       
\usepackage{nicefrac}       
\usepackage{xcolor}         
\usepackage{float}
\usepackage{multirow}
\usepackage{amsthm}
\usepackage{xcolor}
\usepackage{comment}
\usepackage{mathtools}
\usepackage{amssymb}

\usepackage{todonotes}
\usepackage{colortbl}
\usepackage{enumitem}

\title{
Context-Aware Assistant Selection for Improved Inference Acceleration with Large Language Models
}

\author{%
    Jerry Huang\textsuperscript{$\spadesuit\heartsuit$}\thanks{Corresponding authors at: \texttt{jerry.huang@mila.quebec}, \texttt{pp1403@gmail.com}}\quad Prasanna Parthasarathi\textsuperscript{$\clubsuit$}\footnotemark[1] \quad
    \textbf{Mehdi Rezagholizadeh\textsuperscript{$\clubsuit$}\quad Sarath Chandar\textsuperscript{$\spadesuit\diamondsuit\bigstar$}} \\
    \textsuperscript{$\spadesuit$}Chandar Research Lab\quad \textsuperscript{$\spadesuit$}Mila \quad\textsuperscript{$\heartsuit$}Universit\'{e} de Montr\'{e}al \\
    \textsuperscript{$\clubsuit$}Noah's Ark Lab\quad\textsuperscript{$\diamondsuit$}Polytechnique Montr\'{e}al\quad\textsuperscript{$\bigstar$}Canada CIFAR AI Chair
}

\begin{document}

\maketitle

\begin{abstract}
    Despite their widespread adoption, large language models (LLMs) remain prohibitive to use under resource constraints, with their ever growing sizes only increasing the barrier for use. 
    One noted issue is the high latency associated with auto-regressive generation, rendering large LLMs use dependent on advanced computing infrastructure. Assisted decoding, where a \textbf{smaller} \textit{draft} model guides a \textbf{larger} \textit{target} model's generation, has helped alleviate this, but remains dependent on alignment between the two models. Thus if the draft model is insufficiently capable on some domain relative to the target model, performance can degrade. Alternatively, one can leverage multiple draft models to better cover the expertise of the target, but when multiple black-box draft models are available, selecting an assistant without details about its construction can be difficult. To better understand this decision making problem, we observe it as a contextual bandit, where a policy must choose a draft model based on a context. We show that even without prior knowledge of the draft models, creating an offline dataset from only outputs of independent draft/target models and training a policy over the alignment of these outputs can accelerate performance on multiple domains provided the candidates are effective. Further results show this to hold on various settings with multiple assisted decoding candidates, highlighting its flexibility and the advantageous role that such decision making can play.
\end{abstract}

\section{Introduction}

With the introduction of the Transformer~\citep{vaswami2017attention} has emerged the era of large language models~\citep{chowdhery2022palm, touvron2023llama, openai2024gpt4} and the development of LLMs capable of reasoning and acting in astonishingly human-like manner~\citep{kaplan2020scaling, wei2023chainofthought, Ouyang2022TrainingLM}. However, the use of resource intensive models and techniques remains a pre-requisite and accordingly, methods have been developed and applied to alleviate concerns relating to the practical usability of these models~\citep{dettmers2022gptint, dao2024flashattention}. One major area that has observed consistent improvement over time is the \textit{auto-regressive decoding} aspect of text generation, where each generation of a new token requires a complete inference pass through the model, which under-utilizes the property of attention and the ability of modern accelerators (e.g. GPUs, TPUs) to parallelize computations~\citep{deJong2022FiDOFO, kim2023full}.

\begin{figure*}[t!]
    \centering
    \resizebox{0.9\linewidth}{!}{
        \includegraphics[width=\linewidth]{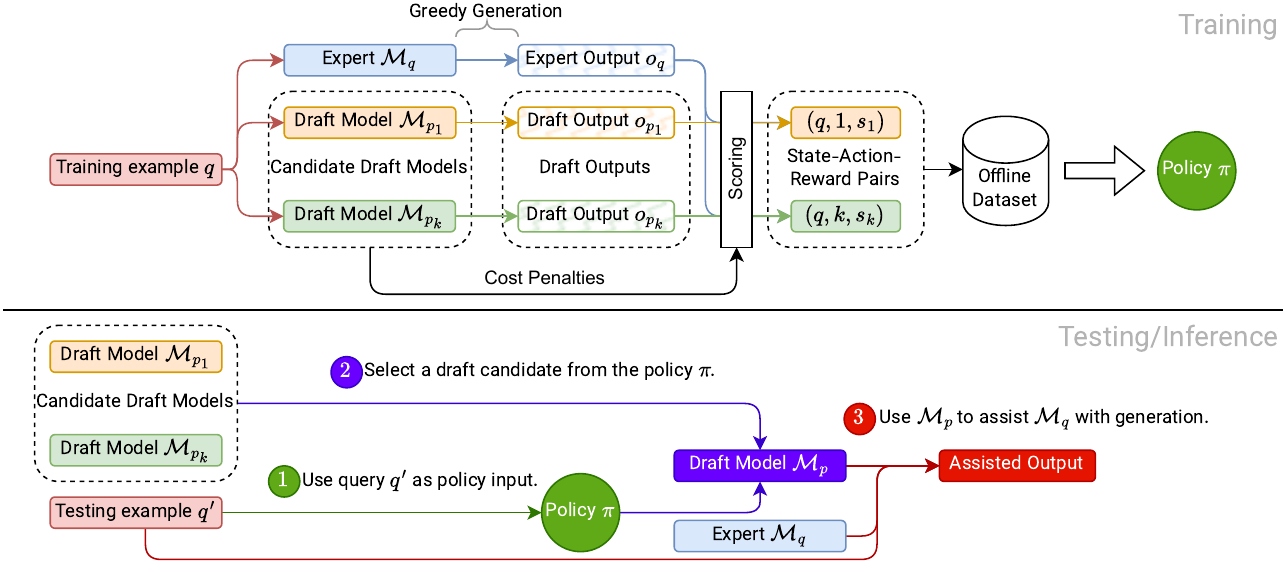}
    }
    \caption{Overview of our methodology. We first train a policy using offline data collected from greedily decoded output from each model, which are scored to produce reward samples. At test time, the policy takes in a query $q'$ to select a draft candidate model, which is then used for assisted generation with the target model.}
    \label{fig:method}
\end{figure*}

A growing approach towards addressing this is speculative decoding~\citep{xia2023speculative, leviathan2023speculativedecoding}.
In speculative decoding, latency is reduced by minimizing the amount of high-latency sequential computations and replacing them with cheaper ones. Rather than sampling directly from the larger model, the sampling is approximated with samples from a smaller and cheaper model through {\it accept-reject} sampling. Specifically, a small draft model auto-regressively generates text which is then verified by a larger target model in parallel~\citep{stern2018blockwise, sun-etal-2021-instantaneous}. Thus the large model does not need to generate text repeatedly but rather guides the small model by correcting outputs when it is truly incapable.
This can reduce the number of calls to the large LLM, saving both time and memory. However two models are required, along with some similarity in their generative abilities in order for this method to see signficant speedups. While approaches exist to circumvent some of these needs~\citep{yang2023predictive, zhang2023draft, li2024eagle, cai2024medusa, hooper2024speed}, these are often limited by the need for additional tuning~\citep{liu2024online, cai2024medusa, li2024eagle}, which is difficult in resource constrained settings, or quality degradation in generations~\citep{kim2023speculative}.
Because of the evident size-cost tradeoff, this is very efficient \emph{if the draft model is well aligned to the target}.

However, while one can ensure that the final output follows the target distribution~\citep{chen2023accelerating}, selecting an inadequate draft model can lead to a lack of acceleration due to the signficant number of rejections that will occur. While other methods allow for changes in the output distribution shift~\citep{kim2023full, zhou2024distillspec, fu2024break} to further speed-up inference, such types of shifts can be problematic in many high-risk scenarios. From this perspective, the presence of multiple draft models, each suited for different settings, can be helpful for inference acceleration without degredations in generation quality. By dynamically choosing a draft model, speedups can be achieved on multiple domains with marginal additional costs. However this requires learning how to choose the best draft option given a context, introducing a decision making problem which needs to be solved.

So how can this decision making process be learned? We start by observing this as a contextual bandits~\citep{contextual-bandits, auer2003}, where the goal is to have a policy select a draft model based on a given query. This also requires rewards from which the policy can learn to estimate and compare the ideality of different actions that can be taken. To this end, we use an offline 
process to estimate the contextual alignment between different draft models and the target model on a set of training examples (\autoref{fig:method}), enabling the construction of a dataset that defines the preference a target can have towards specific draft candidates. This enables us to train a policy that can take into account such preferences without knowing further details about the draft models. By deloying this policy at inference time it becomes possible to weigh these preferences, leading to speedups in generation with the target. We further show that this policy is useful when using self-speculative decoding, whereby the draft model is a subset of the target model parameters. To summarize our contributions:

\begin{itemize}[leftmargin=15pt, itemsep=2pt, parsep=0pt]
    \item We frame a speculative decoding scenario as a contextual bandits problem, where multiple draft models serve as arms that each produce a reward, an abstraction of the inference speed-up relative to using the target model on its own which is not known a priori.
    \item We demonstrate that offline training of a decision making agent through only the similarity between the draft and target model generations, the agent can correctly select which draft model to use for a given input query.
    \item We show that the policy can balance tradeoffs in draft model alignment and generation speed by incorporating explicit information about the model within the reward.
\end{itemize}

\section{Methodology} \label{sec:methodoology}

\subsection{Motivation} \label{subsec:motivation} 

Assume a large target model, $\mathcal{M}_e$, incurs large end-to-end latencies that one wants to avoid. Speculative decoding aims to solve the latency issue by using a draft model to approximate the target model. However, as previously discussed, the draft model must be similar to the target model otherwise the sampling distribution is too different and produce no speedups. Therefore, while draft models can help, they are only reliable when their knowledge distribution resembles that of the target. Accordingly, using only one draft model may not serve well in general if the target has multiple expertises. But by dynamically choosing between different draft models in any given scenario, then benefits from each draft model can be observed as long as the decision maker is competent and efficient.

\subsection{Problem Formulation} \label{subsec:problem}
When presented with a query $q$, selecting a draft model among multiple unique candidates can lead to varying performance based on the chosen option.

From a contextual bandits lens, $q$ is a context for which there are $k$ arms that each returns an independent reward $r$. Each of arm corresponds to a different drafter whose reward is the time it takes to generate the output sequence through speculative decoding. Accordingly, each arm can produce a different reward for each $q$. The objective then consists of learning a policy $\pi(\cdot|q)$ which, for any given context $q$, can select among the arm which can produce the greatest reward. From a speculative decoding scenario, the goal is to select the draft model whose abilities best align with the target for that given query, as this will minimize the number of times the target model must be invoked.

Randomly choosing a draft model risks significant increases in latency, therefore learning to make the correct decision in a sample efficient manner is important. While the ideal reward is the real/observed speed-up, this can be expensive if the aligment with draft models is unknown. As such, a cheaper proxy may be necessary. However, two factors have a direct effect on the true reward: 1) the alignment between target and drafter and 2) the size of the drafter. This provides an alternative way to collect policy training data: use the draft models auto-regressively and compute alignment scores with the target outputs, then adjust these based on the size of the drafter. Next, we describe how we collect our data to train a policy offline.

\subsection{Offline Data Collection} \label{subsec:data}

Given a set of queries $\mathcal{Q} = \{q_i\}_{i=1}^n$, we produce outputs based on each $q_i$ for the target model, $o^e_i$, as well as each of the candidate drafters, $\{o^j_i\}_{j=1}^k$. We then use a similarity metric to compute scores for each candidate output
\begin{equation}
    s^j_i = f(o^e_i, o^j_i)
\end{equation}
as a way to measure the alignment between target and candidates for $q_i$. It is further possible to incorporate a score for the inference speed. For example, if we consider some relative measure of the inference speed for the specific drafter to be $c^j_i$, then one can adjust the score as a weighted sum
\begin{equation}
    s^j_i = \alpha\cdot f(o^e_i, o^j_i) + (1-\alpha) \cdot c^j_i
\end{equation}
which takes into account both factors where $\alpha \in [0, 1]$ weighs the two components.

\subsection{Decision Making} \label{subsec:policy}

With the offline dataset, it becomes possible to train a policy $\pi$ which can independently act on a context by choosing a drafter to use with the target. We consider each $(q_i, j, s^j_i)$ as state-action-reward tuples used to train $\pi$.

Within the contextual bandits reformulation, each query-action pair $(q_t, a_t) \in \mathcal{Q}\times\mathcal{A}(q_t)$ is the drafter which produced an observed reward $r(q_t, a_t)$. Here, we use the score $s^j_i$ directly as the reward, as it acts as an estimate for the effectiveness of drafter $j$ on the context. The policy is represented by a mapping $\pi_\theta(a|q)$ from $\mathcal{Q}\times\mathcal{A}$ to $\mathbb{R}$ and we want to find parameters $\theta^*$ that maximize
\begin{equation*}
    J^\pi = \mathbb{E}_{q\sim P_q(\cdot), a\sim \pi(\cdot|q)}[r(q, a)]
    \label{eq:value_function}
\end{equation*}
where $P_q$ is the sampling distribution of the context. As the action space is discrete, integrating over the action space is equivalent to a
\begin{equation*}
    \int_a \pi_\theta(a|q)da = \sum_{a\in\mathcal{A}(q)} \pi_\theta(a|q) = 1
\end{equation*}
and the gradient with respect to the policy is
\begin{equation*}
    \nabla_\theta J^{\pi_\theta} = \underset{q\sim P_q(\cdot),a\sim \pi_\theta(\cdot|q)}{\mathbb{E}}[\nabla\log\pi_\theta(a|q) r(q, a)]
    \label{eq:value_gradient}
\end{equation*}
which is equivalent to the REINFORCE~\citep{williams2004SimpleSG} policy gradients method and we therefore use it to train our policy.

\section{Experimental Results}

\subsection{Experimental Setup}\label{subsec:exp_setup}
\begin{table*}[ht]

    \centering
    \resizebox{0.95\linewidth}{!}{
        \begin{tabular}{l|ccc|ccc|c}
        \toprule
        \textsc{Draft Model} & \multicolumn{3}{c|}{\textsc{IWSTL2017 En-De}} & \multicolumn{3}{c|}{\textsc{XSum}} & \textsc{Average} \\
        & \textsc{Bleu} & Decoding Speedup & Accept (\%) & \textsc{Rouge-L} & Decoding Speedup &  Accept (\%) & \textsc{Speedup}\\
        \midrule
        \multicolumn{8}{c}{\textbf{Auto-Regressive Generation}} \\
        \midrule
        Greedy & $18.26$ & $1.00\times$ ($31.20_{\pm 0.04}$ ms/token) & - & $35.93$  &  $1.00\times$ ($36.92_{\pm 0.08}$ ms/token) & - & $1.00\times$ \\
        Sampling & $9.85$ & $1.00\times$ ($31.06_{\pm 0.06}$ ms/token) & - & $29.83$ & $1.00\times$ ($37.06_{\pm 0.06}$ ms/token) & - & $1.00\times$ \\
        \midrule
        \multicolumn{8}{c}{\textbf{Greedy Assisted Decoding}} \\
        \midrule
        \texttt{T5-Small} & $18.26$ & $1.10\times$ ($28.24_{\pm 0.15}$ ms/token) & $41.68$ & $35.93$ & $0.97\times$ ($38.60_{\pm 0.18}$ ms/token) & $28.21$ & $1.03\times$ \\
        \texttt{T5-Small-XSum} & $18.26$ & $0.83\times$ ($37.61_{\pm 0.19}$ ms/token) & $7.23$ & $35.93$ & $1.21\times$ ($30.71_{\pm 0.15}$ ms/token) & $40.24$ & $1.02\times$  \\
        \midrule
        \multicolumn{8}{c}{\textbf{Speculative Decoding}} \\
        \midrule
        \texttt{T5-Small} & $9.24$ & $1.10\times$ ($28.14_{\pm 0.17}$ ms/token) & $38.21$ & $29.10$ & $0.99\times$ ($37.53_{\pm 0.14}$ ms/token) & $26.02$ & $1.04\times$ \\
        \texttt{T5-Small-XSum} & $9.51$ & $0.83\times$ ($37.45_{\pm 0.18}$ ms/token) & $8.75$ & $29.10$ & $1.18\times$ ($31.33_{\pm 0.16}$ ms/token) & $38.29$ & $1.01\times$ \\
        \midrule
        \multicolumn{8}{c}{\textbf{Speculative Decoding + Decision Making}} \\
        \midrule
        Greedy $\pi_\theta$ & $9.76$ & $1.09\times$ ($28.56_{\pm 0.16}$ ms/token) & $37.72$ & $29.83$ & $1.17\times$ ($31.63_{\pm 0.19}$ ms/token) & $37.76$ & \textbf{$1.13\times$} \\
        Dynamic $\pi_\theta$ & $9.62$ & $1.07\times$ ($29.20_{\pm 0.17}$ ms/token) & $36.64$ & $29.45$ & $1.16\times$ ($31.88_{\pm 0.17}$ms/token) & $37.34$ & \textbf{$1.11\times$} \\
        \bottomrule
        \end{tabular}
    }
    \caption{Quality, decoding speeds and acceptance rates when using a policy for selecting between different draft models of the same size but specialized on different domains. Across the two domains, both a greedy and dynamic policy can accelerate decoding on both domains, with marginal differences between the exact nature of the policy. Acceptance rate computation is described in \autoref{app:experiment_details}.}
    \label{tab:same_size-diff_domain}
\end{table*}


\paragraph{Models and Tasks.} We select publicly available LLMs to use for our experiments. We conduct a number of experiments, which we motivate by varying the draft options along different axes such as alignment with the target model, sizes of the draft models, architecture of the drafter/target and the level of independence between the draft and target models. Each of these forms a dedicated experiment detailed in the sections that follow.

\paragraph{Data Collection.} For each experiment, we collect offline data using task-specific training dataset splits. Each model is used to generate a greedy decoded sample from each, which is used to construct a reward dataset. To score samples against the target model output, we use the \textsc{Rouge-L} score.

\paragraph{Policy Training.} To train our policy, the input is a sentence embedding of the query from the target model and the output is a distribution over the drafting candidates. We train on the offline dataset for 3 epochs using a fixed batch size of 64 and AdamW~\citep{loshchilov2018decoupled} with a learning rate of 1e-3 and weight decay 1e-2. All other hyperparameters are set to their default values in PyTorch. In all experiments, our policy consists of a 3 layer multi-layer perceptron. Hidden layers have a fixed dimensions of 512 with a \texttt{tanh} activation function. The input dimension is the hidden dimension size of the target model and the output size is the number of drafting options.

\paragraph{Inference.} We sample using a temperature $T$ of 1 and draft tokens $\gamma$ set at 7. We use both a policy that takes the greedy action and another that samples from the output distribution.\footnote{Some ablations are presented in \autoref{app:sampling_methods}.} The policy takes in a sentence embedding of the query and returns a distribution, from which a drafter is sampled and used to assist decoding for the specific query.

\subsection{Results}\label{subsec:results}

\paragraph{Learning to choose the draft model.} For our first experiment, we use a T5~\citep{raffel2020t5} encoder-decoder models. As the target, we use an instruction-finetuned~\citep{wei2022finetuned} \texttt{Flan-T5-XXL}~\citep{chung2022flan} while our draft candidates are publicly available \texttt{T5-Small} models, one the base version and another fine-tuned on text summarization\footnote{Publicly available checkpoint from~\citep{kim2023speculative}.}. We evaluate on translation (\textsc{IWSLT2017 En-De}~\citep{iwslt2017}) and text summarization (\textsc{XSum}~\citep{narayan2018xsum}).

\autoref{tab:same_size-diff_domain} compares when draft models of the same size vary in their domain of expertise. While each model accelerates generation non-trivially within their knowledge domain (\textsc{En-De} for \texttt{T5-Small} and \textsc{XSum} for \texttt{T5-Small-XSum}), they are largely unhelpful or detrimental when used outside their domain of expertise, as seen with the 1\% slowdown from \texttt{T5-Small} on \textsc{XSum} and a 17\% decrease using \texttt{T5-Small-XSum} on \textsc{En-De}. In comparison, the policy ensures acceleration within both domains with neglibile latency from decision making.

\begin{figure*}[ht!] 
    \centering
    \resizebox{0.95\linewidth}{!}{
        \includegraphics[width=0.325\linewidth]{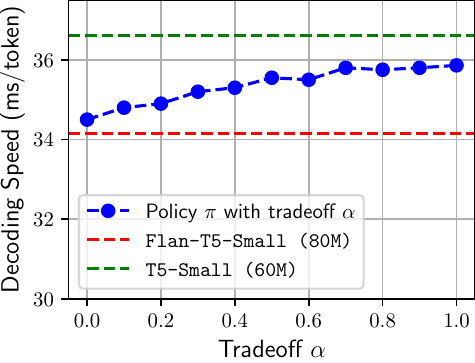}
        \includegraphics[width=0.3\linewidth]{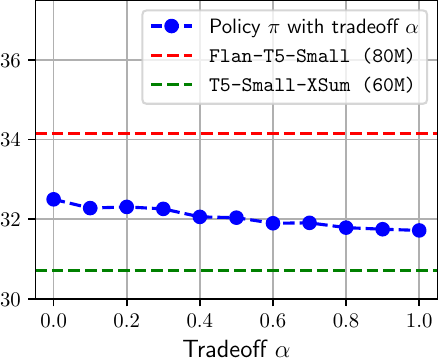}
        \includegraphics[width=0.3\linewidth]{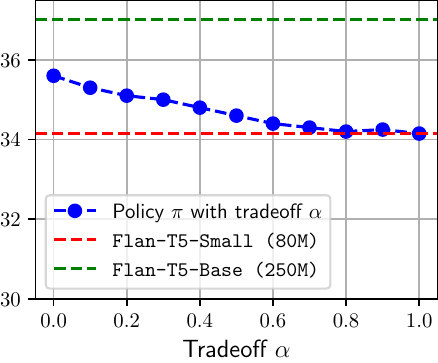}
    }
    \caption{Effect of varying the tradeoff between output alignment and draft model size (controlled through $\alpha$). Each compares the use of \texttt{Flan-T5-Small} as a draft model (red horizontal line). As $\alpha$ increases, the model increasingly uses the smallest draft model for decoding, demonstrating that the offline dataset is sufficient to learn how to balance the quality of the draft model's outputs and the cost of using it. All cases use speculative sampling/decoding.}
    \label{fig:alpha_effect}
\end{figure*}

This highlights some immediate benefits of policy use, namely that it can identify the correct draft model for a context without any explicit information regarding the draft candidates themselves. Rather, generating sampling outputs from each draft model and the target individually is sufficient to develop a general ability to differentiate between domains through the use of the computed rewards.

\begin{table}[ht!]
    \centering
    \resizebox{0.9\linewidth}{!}{
        \begin{tabular}{l|ccccc}
        \toprule
        \textsc{Draft Model} & \textbf{Decoding Speed} \\
        \midrule
        Auto-regressive & $1.00\times$ ($37.06_{\pm 0.16}$ ms/token) \\
        \texttt{T5-Small} & $0.97\times$ ($38.60_{\pm 0.18}$ ms/token) \\
        \texttt{T5-Small-XSum} & $1.21\times$ ($30.71_{\pm 0.15}$ ms/token) \\
        \texttt{Flan-T5-Small} & $1.09\times$ ($34.15_{\pm 0.17}$ ms/token) \\
        \texttt{Flan-T5-Base} & $1.00\times$ ($37.02_{\pm 0.18}$ ms/token) \\
        \bottomrule
        \end{tabular}
    }
    \caption{Speeds of different draft models on \textsc{XSum} with a \texttt{Flan-T5-XXL} model expert (averaged over 5 seeds). Observed decoding speed varies as an effect of drafter size and alignment with the expert.}
    \label{tab:diff_size-same_domain}
\end{table}

\begin{figure*}[ht!]
    \centering
    \resizebox{0.9\linewidth}{!}{
        \includegraphics[width=0.52\linewidth]{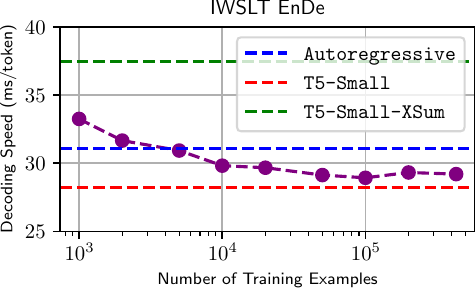}
        \includegraphics[width=0.48\linewidth]{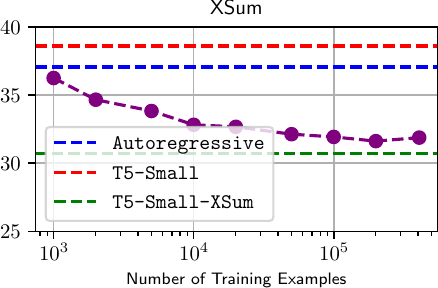}
    }
    \caption{Decoding speed using a dynamic policy as a function of the number of examples used to train the policy, tested on \textsc{IWSLT2017 En-De} (left) and \textsc{XSum} (right). The number of examples is marked in $\log$-scale. Horizontal lines denote decoding speeds of individual drafting options.}
    \label{fig:policy-training-examples}
\end{figure*}

\paragraph{Balancing quality and speed.} It is also important that the draft model is sufficiently inexpensive to use relative to the target model. This motivates our second experiment, which is evaluated only on \textsc{XSum}, but compares draft candidates that vary in terms of size and target model alignment. Multiple draft models are compared: a \texttt{Flan-T5-Small} (80M parameters), the same \texttt{T5-Small} (60M) models mentioned above, and \texttt{Flan-T5-Base} (220M).
\autoref{tab:diff_size-same_domain} shows the speed-ups earned through speculative decoding using these different draft candidates and a \texttt{Flan-T5-XXL} target. Although larger draft models may be better aligned with the target compared to smaller options, using them incurs a latency that can end up being less efficient.

Balancing alignment and efficiency is therefore an issue to consider when deciding between candidates. \autoref{fig:alpha_effect} shows how offline training can help accomplish this. Setting $\alpha$ to vary between the objectives defined in \S\ref{subsec:data}, where we use fixed inference costs based on the size of the draft models, we observe how a dynamic policy can eventually adapt to the preferences set by the choice of $\alpha$. For example, as $\alpha$ approaches 1, the policy places increasing preference on the smallest draft model regardless of quality. Meanwhile $\alpha \rightarrow$ 0 shows increasing preference towards the model that has greatest alignment with the target generations.

This demonstrates the general flexibility that can come with using such a weighting scheme of different rewards, while demonstrating that even simpler proxies for the inference penalty are sufficient to properly balance the two.

\paragraph{How many examples need to be learned to differentiate?} It is further necessary to consider the number of examples that are needed for the decision maker to properly learn to differentiate between different examples. To this end, we investigate how quickly the policy can learn to use the annotated scores within the offline dataset to demonstrate a visible speed-up improvement. We re-use our models from the first experiment, but keep track of the decoding speed as the number of examples used to train our policy $\pi_\theta$ increases.

As we can observe in \autoref{fig:policy-training-examples}, learning to select the correct model occurs rather quickly, as training for fewer than a total of 10000 examples is sufficient to attain a level of performance that is equivalent to training on the entire offline dataset, which consists of nearly 400 thousand examples. This result demonstrates the general efficiency of this method, as collecting and training the policy on outputs from a minimal amount of examples shows the ability to generalize quite strongly.

\paragraph{Auto-regressive generation as an option.} Scenarios exist where the draft models will not be useful, in which case using the target auto-regressively remains the most reasonable option.

To this end, we attempt to observe how providing this option to the decision maker can affect our previous experiments. We repeat the same experiment from \autoref{tab:same_size-diff_domain} but allow our policy to learn to choose to generate auto-regressively. To avoid trivially perfect matching of outputs, we sample outputs from the target model and score against the greedy output. Due to the large size of the target compared to the drafters, we use $\alpha=0.5$ to balance the size and quality scores.

\begin{table}[ht!]
    \centering
    \resizebox{1.0\linewidth}{!}{
        \begin{tabular}{l|cc}
        \toprule
        \textbf{\textsc{En-De}} & \textbf{\textsc{Bleu}} & \textbf{Decoding Speed} \\
        \midrule
        Greedy $\pi_\theta$ & $9.82$ & $1.09\times$ ($28.95_{\pm 0.15}$ ms/token) \\
        Dynamic $\pi_\theta$ & $9.76$ & $1.07\times$ ($29.34_{\pm 0.13}$ ms/token) \\ 
        \bottomrule
        \toprule
        \textbf{\textsc{XSum} }& \textbf{\textsc{Rouge-L}} & \textbf{Decoding Speed} \\
        \midrule
        Greedy $\pi_\theta$ & $30.02$ & $1.17\times$ ($31.64_{\pm 0.17}$ ms/token) \\
        Dynamic $\pi_\theta$ & $29.53$ & $1.16\times$ ($31.73_{\pm 0.18}$ ms/token) \\
        \bottomrule
        \end{tabular}
    }
    \caption{
        Decoding speeds under a dynamic decision making regime where auto-regressive generation is a decoding option, on \textsc{IWSLT2017 En-De} and \textsc{XSum}. 
    }
    \label{tab:with_autoregressive}
\end{table}

\begin{table}[ht!]
    \centering
    \resizebox{\linewidth}{!}{
        \begin{tabular}{l|cc}
        \toprule
        \textsc{Draft Candidate} & \textbf{\textsc{Accuracy}} & \textbf{Decoding Speed} \\
        \midrule
        Auto-Regressive & $10.67$ & $1.00\times$ ($22.03_{\pm 0.16}$ ms/token) \\
        \texttt{T5-Small} & $10.99$ & $0.78\times$ ($28.08_{\pm 0.18}$ ms/token) \\
        \texttt{T5-Small-XSum} & $10.09$ & $0.74\times$ ($29.91_{\pm 0.16}$ ms/token) \\
        \midrule
        Greedy $\pi_\theta$ & $10.64$ & $0.95\times$ ($23.54_{\pm 0.19}$ ms/token) \\
        Dynamic $\pi_\theta$ & $10.38$ & $0.95\times$ ($23.45_{\pm 0.18}$ ms/token) \\
        \bottomrule
        \end{tabular}
    }
    \caption{Decoding speeds on \textsc{GSM8K} (test set) with a \texttt{Flan-T5-XXL} expert. 
    Inference on the latter two tasks is negligibly different from \autoref{tab:with_autoregressive}.}
    \label{tab:gsm8k}
\end{table}

\autoref{tab:with_autoregressive} shows that adding the auto-regressive option does not degrade decoding speed or generation quality and in fact may accelerate overall decoding. This may be because some contexts are out-of-distribution for all candidate draft model; in these cases, auto-regressive generation can be much more efficient by a signficant margin. 

We further verify whether the policy can ignore draft models when they are not useful. We experiment by including \textsc{GSM8K} to our tasks, which only the target is aligned. Since neither draft model can accelerate inference on this task, the policy should ideally avoid drafting for examples from this setting. Since \textsc{GSM8K} is significantly smaller than \textsc{En-De} and \textsc{XSum}, we reduce the number of examples to match all datasets in terms of size.

\autoref{tab:gsm8k} shows auto-regressive generation to (unsurprisingly) outperform assisted generation. However, using a policy shows comparable speed to auto-regressive generation, indicating that it learns to ignore the draft models due to the stark contrast in the greedy outputs from each model.

\begin{table*}[ht!]
    \centering
    \resizebox{0.95\linewidth}{!}{
        \begin{tabular}{l|cccccc|c}
            \toprule
            Method & \textsc{MT-Bench} & \textsc{Trans.} & \textsc{Sum.} & \textsc{QA} & \textsc{Math} & \textsc{RAG} & \textsc{Avg.} \\
            \midrule
            Auto-regressive & $1.00\times$ & $1.00\times$ & $1.00\times$ & $1.00\times$ & $1.00\times$ & $1.00\times$ & $1.00\times$ \\
            \texttt{Vicuna-68m} & $1.74\times$ & $1.28\times$ & $1.74\times$ & $1.54\times$ & $1.70\times$ & $1.63\times$ & $1.60\times$ \\
            \texttt{Vicuna-160m} & $1.52\times$ & $1.10\times$ & $1.51\times$ & $1.36\times$ & $1.50\times$ & $1.44\times$ & $1.40\times$ \\
            \texttt{LLaMA-68m} & $1.40\times$ & $1.32\times$ & $1.41\times$ & $1.41\times$ & $1.48\times$ & $1.49\times$ & $1.41\times$ \\
            \midrule
            Policy $\pi_\theta$ + \texttt{Vicuna-68m/160m} & $1.74\times$ & $1.28\times$ & $1.73\times$ & $1.54\times$ & $1.70\times$ & $1.62\times$ & $1.59\times$ \\
            Policy $\pi_\theta$ + \texttt{Vicuna-68m/LLaMA-68m} & $1.72\times$ & $1.31\times$ & $1.73\times$ & $1.53\times$ & $1.69\times$ & $1.63\times$ & $1.60\times$ \\
            Policy $\pi_\theta$ + \texttt{Vicuna-160m} and \texttt{LLaMA-68m}  & $1.51\times$ & $1.31\times$ & $1.49\times$ & $1.39\times$ & $1.49\times$ & $1.47\times$ & $1.41\times$ \\
            \bottomrule
        \end{tabular}
    }
    \caption{Acceleration on SpecBench using a \texttt{Vicuna-33B} expert with \texttt{Vicuna-68m}, \texttt{Vicuna-160m} and \texttt{LLaMA-68m} as draft candidates.
    We use a greedy policy to select a drafter for a given query.}
    \label{tab:specbench}
\end{table*}

\paragraph{Generalization to Multi-Task Drafters.} To demonstrate the applicability of this method to more general settings, in particular cases where the draft models may be competent at multiple tasks, we further apply our policy-based selection method to SpecBench~\citep{xia2024unlocking} using a \texttt{Vicuna-33B}~\citep{vicuna2023} target with smaller draft models (\autoref{tab:specbench}). Given the size of SpecBench (480 examples, divided equally into 6 tasks), we use this exclusively as a test-set. To train our policy, we use the original task datasets from which SpecBench examples were extracted and sample even amounts of examples from each (2000). For \textsc{MT-Bench}, there are only 80 total examples which are all included in the test set but which we sample with replacement to use for a training set. Accordingly, results on this task may be over-confident. Because \texttt{Vicuna} models are decoder-only Transformers, we adjust the sentence representation to be the final hidden representation of the input sequence. Our results show that our initial findings from a \texttt{T5} architecture hold, suggesting that such a policy-based training method is both robust and generalizable to different settings.

\paragraph{Ablation with self-drafting.}
Despite the benefits of assisted decoding, drafting relies on the availability of small draft models that
\begin{enumerate}[label=(\arabic*), itemsep=2pt, parsep=0pt]
    \item Share a vocabulary with the target.
    \item Align with the target on the tasks of interest.
\end{enumerate}
Such models can be difficult to obtain, leading to \textit{self-drafting}~\citep{yang2023predictive, li2024eagle, hooper2024speed}, where the draft model exists within the target. To explore the differences with this setting, we conduct an additional ablation.

We use a \texttt{LLaMA-2-13B-Chat} model~\citep{touvron2023llama} using early exits, following \citep{kavehzadeh2024sorted} with the use of a single language modeling head for all exits. While other methods exist, these can possess a combinatorial number of potential draft options and necessitate pre-determined path flows during inference~\citep{zhang2023draft}. Meanwhile, methods that use additional language modeling heads for parallel decoding require additional parameters which can both become irreconcilable with resource constraints or degrade generation quality~\citep{cai2024medusa}.
\begin{table}[t!]
    \resizebox{\linewidth}{!}{
        \centering
        \begin{tabular}{l|cc}
        \toprule
        \textsc{Draft Candidate} & \textbf{\textsc{Alpaca}} & \textbf{\textsc{TriviaQA}} \\
        \midrule
        \multicolumn{3}{c}{\textbf{Auto-Regressive Generation}} \\
        \midrule
        Auto-regressive & $1.00\times$ ($26.43_{\pm 0.14}$ ms/token) & $1.00\times$ ($6.12_{\pm 0.12}$ ms/token) \\
        \midrule
        \multicolumn{3}{c}{\textbf{Intermediate Drafting}} \\
        \midrule
        Layer 8 Drafting & $0.77\times$ ($34.24_{\pm 0.15}$ ms/token) & $0.77\times$ ($7.97_{\pm 0.14}$ ms/token) \\
        Layer 16 Drafting & $0.81\times$ ($32.51_{\pm 0.13}$ ms/token) & $0.69\times$ ($8.87_{\pm 0.15}$ ms/token) \\
        Layer 24 Drafting & $0.73\times$ ($36.07_{\pm 0.16}$ ms/token) & $0.69\times$ ($8.93_{\pm 0.13}$ ms/token) \\
        Layer 32 Drafting & $0.50\times$ ($52.82_{\pm 0.19}$ ms/token) & $0.60\times$ ($10.21_{\pm 0.13}$ ms/token) \\
        \midrule
        \multicolumn{3}{c}{\textbf{Intermediate Drafting + Policy}} \\
        \midrule
        Greedy Policy $\pi_\theta$ & $0.80\times$ ($32.85_{\pm 0.16}$ ms/token) & $0.66\times$ ($9.41_{\pm 0.14}$ ms/token) \\
        Dynamic Policy $\pi_\theta$ & $0.82\times$ ($32.17_{\pm 0.15}$ ms/token) & $0.66\times$ ($9.34_{\pm 0.11}$ ms/token) \\
        \midrule
        \multicolumn{3}{c}{\textbf{Intermediate Drafting + Policy + Auto-regressive Option}} \\
        \midrule
        Greedy Policy $\pi_\theta$ & $0.99\times$ ($26.75_{\pm 0.18}$ ms/token) & $0.96\times$ ($6.47_{\pm 0.12}$ ms/token) \\
        Dynamic Policy $\pi_\theta$ & $0.95\times$ ($27.58_{\pm 0.19}$ ms/token) & $0.91\times$ ($6.78_{\pm 0.13}$ ms/token) \\
        \bottomrule
        \end{tabular}
     }
   \caption{Results in a scenario for deciding when to early exit. 
    {Even using a self-drafting model with multiple early exists, the policy is capable of maintaining performance by choosing an appropriate action.}
    }
    \label{tab:sorted}
\end{table}
Results on \textsc{Alpaca} and \textsc{TriviaQA}, conducted on each dataset independently, under this setup (\autoref{tab:sorted}) show that although intermediate layer drafting results in a decrease in decoding speed, using a policy can minimize performance loss in particular with the presence of an auto-regressive option, highlighting that the proposed offline policy learning approach has potential for self-drafting as well. This demonstrates the use of a policy remains a useful manner to fall back to the most effective decoding options. Furthermore, when considering the case where the auto-regressive option is not available, we note that the policy methods are capable of recovering to a performance similar to the best case intermediate drafter on \textsc{Alpaca}. While this is not the case with \textsc{TriviaQA}, this is perhaps attributed to the short answers within this dataset.

\section{Discussion}

\paragraph{LLM Routing.} LLMs have demonstrated remarkable capabilities across a range of tasks, but there exists wide variation in their costs and capabilities. Very broadly, more capable models tend to be more expensive than less capable models. This leads to a dilemma when deploying LLMs in the real-world - routing all queries to the largest, most capable model leads to the highest-quality responses but can be expensive, while routing queries to smaller models can save costs but may result in lower-quality responses. Similarly, not all models may be well suited for the same set of tasks, meaning that routing to the most suitable model can be of great importance as well.

Our work shares a great deal of similarity with this notion of model routing, or selecting the best model based on the query. In particular, the set of draft models can be considered to be a group of sub-networks, similar to a Mixture-of-Experts (MoE)~\citep{shazeer2017} style paradigm. The policy meanwhile acts as a router to the correct sub-network. More advanced routing techniques~\cite{Fedus2021SwitchTS, ong2024routellmlearningroutellms} have been explored as a way to leverage the multitude of LLMs that exist in the wild, but have yet to be widely used within downstream settings such as speculative decoding.

\paragraph{Adaptive Speculative Decoding.} Speculative decoding methods require the use of many pre-defined hyper-parameters which can signficantly influence acceleration, with even minor changes having noticable effects. Recent work has begun to explore how to decouple this process, such as by dynamically selecting the number of drafting tokens to generate at each decoding step~\citep{wang2024minions, liu2024speculative}. \citet{kavehzadeh2024sorted} further discussed dynamically selecting a model per instance, however their method is limited to their specific setup due to needing to compute confidence scores after generation at early exits.

While we do not introduce a new decoding algorithm, we make a first attempt to make the speculative decoding adaptive through the ability to switch between multiple draft models based on the input. However, more complex levels of adaptivity may be necessary as each decoding step may not be the same, necessitating perhaps a need to carefully adjust different hyperparameters through the process in order to maximize acceleration.

\paragraph{Decision Making for Assisted Decoding.} Assisted decoding can require making multiple decisions. One of these is determining an ideal number of draft tokens to decode at each step. Another relates to how to reject tokens, which commonly uses either greedy~\citep{xia2023speculative} or sampling-based token-matching heuristics~\citep{leviathan2023speculativedecoding}. However, there are trade-offs when enforcing specific choices, which requires further investigation to better understand how to tune such techniques.

This work proposes adding an additional decision at the beginning of the decoding process, namely at the beginning of the process under the assumption that multiple drafting options exist. While we limit ourselves to make a more complete analysis within a more self-contained setting, various ways to have these methods co-exist within one larger pipeline are possible. However such work is left for future exploration due to the non-trivial nature of understanding how different choices and effect overall reported results in conjunction.

\paragraph{Measuring Alignment Between Outputs.} 

We observe that token-level similarity scores are effective for training the decision maker, which can be attributed to the fact that assisted decoding itself relies on matching the token-level distribution of outputs. As such, if the greedy-decoded output from a draft model highly resembles the target output, it follows that this will be represented by a higher degree of similarity between the probabilty distributions in the logit space, which can then lead to fewer rejections when sampling.

However, such metrics have limitations~\citep{deutsch2022limitations} due to capturing primarily superficial elements of text, where marginal differences in distribution have large effects on the output text. Furthermore, different metrics may overfit specific tasks, necessitating the need for better measures of draft/target alignment, which can hopefully lead to better estimation of rewards for training improved policies, either by desigining better metrics themselves or by learning to compare features at different levels of granularity (ex. target and draft logits against text outputs). Additionally, semantic meaning also can play an important role, as outputs with signficant structure may still possess the same meaning, something that token-level similarity metrics will not adequately capture.

\paragraph{Speculative Decoding as Approximate Inference.} Speculative decoding can be analogized as a form of approximate inference where due to the intractability of performing inference with a model of interest, approximation methods are used to learn an estimate of the model. While training the draft model is equivalent to performing variational inference (i.e. approximating an intractable distribution with a surrogate), this can be expensive. Accordingly, training only a policy can be seen as weighing a set of fixed distributions to act as a better surrogate for the target model.

Some works have further attempted to study speculative decoding from this angle. In particular, \citet{zhou2024distillspec} explore such a process by building a draft model through the use of KL-divergence losses, effectively building a posterior distribution of the target model based on likelihood information from the draft output. \citet{liu2024online} meanwhile explore the same technique as the distribution of examples changes, building a draft model that can adapt to changing user inputs. Such settings also could perhaps benefit from multiple draft models, where conditioning on the query can enable more effective adaptation of draft models to better generalize to unseen settings.

\paragraph{Hosting Multiple Draft Models.} An important aspect of this method relates to the need to host multiple draft models in conjunction with the expert. This can incurr additional costs, in particular if the expert and selected drafter do not reside in the same device. While methods such as self-drafting avoid this issue and the possibility to create minimally-sized drafters generally alleviates the concern of excessive memory usage, one particular aspect of consideration remains hardware level optimizations which can best enable for the selected drafters to be loaded at maximal speed, avoiding additional latency that can result from the bandwidth constraints that relate to data transfer between devices.

\section{Conclusion}

This work presents the first work at attempting to integrate assisted generation within a setting where multiple black-box draft candidates exist. When no a-priori knowledge of which draft candidate is best suited for assisting the decoding of a given example, the problem can be modeled as a contextual bandits problem, where the goal is to estimate the unknown reward from each drafting option. Our work demonstrates that offline RL presents an efficient method for learning to distinguish the available options and provide accelerated decoding across various examples within this setting, with a logical way to collect offline data from models for learning. Our results and ablations show that learning a policy with this approach can adapt to general preferences while accounting for more complex aspects of the decision making, highlighting its robustness. Furthermore, such a method is scalable and robust to the introduction of more draft models or the removal of draft models, presenting a viable alternative to settings where a uniquely superior draft model may be unavailable.

Nevertheless, areas of further development exist. For example, learning in an online fashion may render this method more broadly applicable. Alternatively, exploring how to dynamically choose drafters at every decoding step rather than per example, as well as combining this direction of work with that which attempts to adaptively choose the speculation length at every step, are feasible ways of combining our findings with concurrent work in the hopes of reaping the benefits of all methods.

\section{Limitations}

This work has a few limitations which define the scope of future work.

\subsection*{Choice of draft models and data domains} Results may stem from the distinct boundaries that exist between domains/tasks. In settings where such boundaries are not well defined, outcomes may differ. However technical limitations and the absence of sufficent pre-trained models for comparision makes this difficult to explore immediately.

\subsection*{Additional storage and memory} 
The usage of multiple models that draft independently requires additional memory, which can be be more difficult to manage when there are explicit constraints on this front (self-drafting avoids this due to the use of a single model). Furthermore, collecting an offline dataset can be difficult in some specific scenarios where inference is burdensome, for example when input/output sequences are very long, or when many offline examples are required.

\subsection*{Self-Drafting}
We work on a setting where we do not conduct any additional training of parameters that are explicitly linked to the language model itself, whether they are existing parameters or new paramters added as a result of the method. While there are ways in which our explored method can be applied to these as well, computational limitations make it difficult to rigorously conduct such studies at the moment and we leave it to future work for this reason.

\section{Ethics Statement}

This paper discusses the concept of dynamically choosing between of multiple black-box draft models for speculative decoding, proposing an offline reinforcement learning approach for adaptively selecting a good draft model for assistance. Our results are relate to the decoding speed of models, which is unlikely to lead to ethical concerns or problematic interpretations of such results.

\section{Acknowledgements}

Jerry Huang recieved financial support under the form of a National Science and Engineering Research Council (NSERC) Canada Graduate Scholarship, a Fonds de Recherche du Qu\'{e}bec Nature et technologies (FRQNT) Training Scholarship and a Bourse d'Excellence Hydro-Qu\'{e}bec. Sarath Chandar is supported by a Canada CIFAR AI Chair, the Canada Research Chair in Lifelong Machine Learning and a NSERC Discovery Grant.

\bibliography{refs}

\appendix

\section{Experimental Details}\label{app:experiment_details}

\begin{table*}[ht]
    \centering
    \resizebox{\linewidth}{!}{
    \begin{tabular}{cccc|cc|c}
        \toprule
         Target Model & Draft Model & Temperature & Draft Tokens & Ours & Original & Relative Difference \\
         \midrule
         \texttt{T5-XXL} & \texttt{None} & 0 & - & $1.00\times$ (19.6 ms/token) & $1.0\times$ & -\\
         \midrule
         \texttt{T5-XXL} & \texttt{T5-Small} & 0 & 7 & $1.21\times$ (16.2 ms/token) & $3.4\times$ & $2.81\times$\\
         \texttt{T5-XXL} & \texttt{T5-Base} & 0 & 7 & $0.96\times$ (20.4 ms/token) & $2.8\times$ & $2.91\times$\\
         \texttt{T5-XXL} & \texttt{T5-Large} & 0 & 7 & $0.61\times$ (32.1 ms/token) & $1.7\times$ & $2.78\times$\\
         \midrule
         \texttt{T5-XXL} & \texttt{T5-Small} & 1 & 7 & $1.03\times$ (19.1 ms/token) & $2.6\times$ & $2.53\times$\\
         \texttt{T5-XXL} & \texttt{T5-Base} & 1 & 5 & $0.83\times$ (23.5 ms/token) & $2.4\times$ & $2.88\times$\\
         \texttt{T5-XXL} & \texttt{T5-Large} & 1 & 3 & $0.56\times$ (35.3 ms/token) & $1.4\times$ & $2.52\times$\\  
         \bottomrule
    \end{tabular}
    }
    \caption{Reproduced translation results}
    \label{tab:reproduction-translation}
\end{table*}

\begin{table*}[ht]
    \centering
    \resizebox{\linewidth}{!}{
    \begin{tabular}{cccc|cc|c}
        \toprule
         Target Model & Draft Model & Temperature & Draft Tokens & Ours & Original & Relative Difference \\
         \midrule
         \texttt{T5-XXL} & \texttt{None} & 0 & - & $1.00\times$ (31.8 ms/token) & $1.0\times$ & -\\
         \midrule
         \texttt{T5-XXL} & \texttt{T5-Small} & 0 & 5 & $0.99\times$ (32.1 ms/token) & $3.1\times$ & $3.13\times$ \\
         \texttt{T5-XXL} & \texttt{T5-Base} & 0 & 5 & $0.87\times$ (36.4 ms/token) & $3.0\times$ & $3.43\times$ \\
         \texttt{T5-XXL} & \texttt{T5-Large} & 0 & 3 & $0.59\times$ (53.8 ms/token) & $2.2\times$ & $3.72\times$  \\
         \midrule
         \texttt{T5-XXL} & \texttt{T5-Small} & 1 & 5 & $0.86\times$ (33.0 ms/token) & $2.3\times$ & $2.39\times$ \\
         \texttt{T5-XXL} & \texttt{T5-Base} & 1 & 5 & $0.85\times$ (37.5 ms/token) & $2.2\times$ & $2.59\times$ \\
         \texttt{T5-XXL} & \texttt{T5-Large} & 1 & 3 & $0.57\times$ (48.6 ms/token) & $1.7\times$ & $2.60\times$ \\  
         \bottomrule
    \end{tabular}
    }
    \caption{Reproduced summarization results}
    \label{tab:reproduction-summarization}
\end{table*}

\subsection{Baselines}

To baseline and compare our architectural constraints against \citep{leviathan2023speculativedecoding}, we partially benchmark our experiments against theirs. These are presented in \autoref{tab:reproduction-translation} and \ref{tab:reproduction-summarization}. We conduct this as we suspect a difference in both system architechture used for experiments as well as for implementation of models.

We observe that their results are generally show speedups that are consistently 2.5 to 3.0$\times$ as large as ours, with minor deviations. We attribute this to the usage of different computational resources and potential implementation differences. Additionally, given the small amount of variation in the relative differences between the observed and reported speedups, we contend that these differences are not due to errors in implementation.

\subsection{Technical Details}\label{app:technical_details}

All experiments are conducted on a machine with a single NVIDIA A100 GPU with 8 CPU cores. We run all experiments using PyTorch and HuggingFace models.

\subsection{Hyperparameter Configurations}\label{app:hyperparams}

Details about hyperparameters we use within our experiments are detailed here.

\begin{table}[ht!]
    \centering
    \begin{tabular}{lr}
    \toprule
    \textbf{Hyperparamter} & \textbf{Value} \\
    \midrule
    Optimizer     & AdamW \\
    Learning Rate & 0.001 \\
    Weight Decay  & 0.01 \\
    $\beta_1$     & 0.9 \\
    $\beta_2$     & 0.99 \\
    $\epsilon$    & 1e-8 \\
    \bottomrule
    \end{tabular}
    \caption{Optimization Hyperparameters}
    \label{tab:hyperparameters}
\end{table}

\subsection{Templates}

For each example, we use a specific prompt based on the dataset from which the data originates (see \autoref{tab:prompts}). We follow the templates provided originally by \citet{raffel2020t5} and \citet{chung2022flan}.

\begin{table}[ht!]
    \centering
    \resizebox{\linewidth}{!}{
        \begin{tabular}{lr}
        \toprule
        \textbf{Task} & \textbf{Prompt} \\
        \midrule
        \textsc{EnDe} & \texttt{translate English to German: \{input\}} \\
        \textsc{XSum} & \texttt{summarize: \{input\}} \\
        \textsc{GSM8K} & \texttt{Q: \{input\}} \\
        \bottomrule
        \end{tabular}
    }
    \caption{Prompts for the different tasks}
    \label{tab:prompts}
\end{table}

\subsection{Cost Function}

When considering multiple draft candidates, we use the following simple function for generating fixed costs for the different models. Suppose the candidates have $P = \{p_1, p_2, \dots, p_k\}$ parameters. Then the cost for the models are
$$
c_i = 1-\frac{e^{p_i}}{e^{p_j}}
$$
where $j = \underset{i}{\text{argmax}} \phantom{0} p_i$.

\subsection{Accept Rate Computation}\label{app:accept_rate}

To compute the accept rate of tokens, we define the \textbf{number of generated tokens} in a given draft as the total number of tokens generated by the draft model (this is equivalent to $\gamma$). The \textbf{number of accepted tokens} in a given draft is the number of generated tokens that are validated as correct by the target model. When a token is rejected within a draft, all subsequent tokens are considered rejected as well. The accept rate is then the quotient of the total number of accepted tokens divided by the total number of generated tokens.

\subsection{Computing Wall-Clock Performance}\label{app:clock}

To compute the wall-clock time when using a policy, we include the amount of time used to infer on the policy. However, we do not include the time needed to generate the sentence representation.

This is because upon generating the original sentence representation, the large model's KV cache can be updated to store these for the future verification passes, meaning that they do not need to be recomputed again in the future. As such, we treat this initial pass through as being part of the first verification pass.

Additionally, one could theoretically save on the policy inference by performing batched inference on many examples at once. However, this is not particularly applicable in practice, where different inputs arrive at different times. As such, we treat each example individually and include these times within the per-example speeds.

\section{Different Sampling Hyperparameters}\label{app:sampling_methods}

We run our ablations based on our setup for our first experiment.

\subsection{Effect of Number of Draft Tokens}\label{app:draft_tokens}

We test our method with 5, 7 and 10 draft tokens in \autoref{tab:iwstl-tokens} and \autoref{tab:xsum-tokens}.

\begin{table*}[ht]
    \resizebox{\linewidth}{!}{
        \centering
        \begin{tabular}{l|ccc}
        \toprule
        \textsc{Draft Model} & $\gamma=5$ & $\gamma=7$ & $\gamma=10$ \\
        \hline
        \multicolumn{4}{c}{\cellcolor[gray]{0.8}{\textbf{Speedup}}} \\
        \hline
        Auto-regressive 
            & \multicolumn{3}{c}{\cellcolor[gray]{0.9}{$1.00\times$ (31.06 ms/token)}} \\
        \texttt{T5-Small} 
            & $1.16\times$ (28.93 ms/token) 
            & $1.19\times$ (28.14 ms/token)
            & $1.16\times$ (28.95 ms/token) \\
        \texttt{T5-Small-XSUM} 
            & $0.80\times$ (38.92 ms/token) 
            & $0.83\times$ (37.61 ms/token) 
            & $0.82\times$ (38.04 ms/token) \\
        \midrule
        Greedy Policy $\pi_\theta$ 
            & $1.09\times$ (28.45 ms/token)
            & $1.09\times$ (28.56 ms/token)
            & $1.09\times$ (28.53 ms/token) \\
        Dynamic Policy $\pi_\theta$ 
            & $1.06\times$ (29.53 ms/token)
            & $1.07\times$ (29.20 ms/token)
            & $1.06\times$ (29.77 ms/token) \\
        \bottomrule
        \end{tabular}
    }
    \caption{Varying the number of drafted tokens for assisted generation on \textsc{IWSLT2017 En-De}. $T=1$ for all cases.}
    \label{tab:iwstl-tokens}
\end{table*}

\begin{table*}[ht]
    \resizebox{\linewidth}{!}{
        \centering
        \begin{tabular}{l|ccc}
        \toprule
        \textsc{Draft Model} & $\gamma=5$ & $\gamma=7$ & $\gamma=10$ \\
        \hline
        \multicolumn{4}{c}{\cellcolor[gray]{0.8}{\textbf{Speedup}}} \\
        \hline
        Auto-regressive 
            & \multicolumn{3}{c}{\cellcolor[gray]{0.9}{$1.00\times$ (37.06 ms/token)}} \\
        \texttt{T5-Small} 
            & $0.96\times$ (38.84 ms/token)
            & $0.97\times$ (38.60 ms/token)
            & $0.96\times$ (38.98 ms/token) \\
        \texttt{T5-Small-XSUM} 
            & $1.19\times$ (31.20 ms/token)
            & $1.21\times$ (30.71 ms/token) 
            & $1.19\times$ (31.39 ms/token) \\
        \midrule
        Greedy Policy $\pi_\theta$ 
            & $1.15\times$ (32.30 ms/token)
            & $1.17\times$ (31.63 ms/token)
            & $1.14\times$ (32.44 ms/token) \\
        Dynamic Policy $\pi_\theta$ 
            & $1.15\times$ (32.33 ms/token)
            & $1.16\times$ (31.88 ms/token) 
            & $1.14\times$ (32.50 ms/token) \\
        \bottomrule
        \end{tabular}
    }
    \caption{Varying the number of drafted tokens for assisted generation on \textsc{XSum}. $T=1$ for all cases.}
    \label{tab:xsum-tokens}
\end{table*}

\subsection{Effect of Temperature}\label{app:temperature}
We test our method with varying temperature values in \autoref{tab:iwstl-temp} and \autoref{tab:xsum-temp}.

We noted through ablations that increasing temperature past $T=1$ resulted in a signficant slowdown in the decoding speed as well as the quality of the sampled generations. As such, we only present results on values of $T \leq 1$.

\begin{table*}[ht]
    \resizebox{\linewidth}{!}{
        \centering
        \begin{tabular}{l|ccc}
        \toprule
        \textsc{Draft Model} & $T=0.5$ & $T=0.9$ & $T=1$ \\
        \hline
        \multicolumn{4}{c}{\cellcolor[gray]{0.8}{\textbf{Speedup}}} \\
        \hline
        Auto-regressive 
            & \multicolumn{3}{c}{\cellcolor[gray]{0.9}{$1.00\times$ (31.06 ms/token)}} \\
        \texttt{T5-Small} 
            & $1.06\times$ (29.00 ms/token)
            & $1.09\times$ (28.36 ms/token)
            & $1.10\times$ (28.14 ms/token) \\
        \texttt{T5-Small-XSUM}
            & $0.83\times$ (37.70 ms/token) 
            & $0.81\times$ (37.53 ms/token) 
            & $0.83\times$ (37.61 ms/token) \\
        \midrule
        Greedy Policy $\pi_\theta$ 
            & $1.11\times$ (27.83 ms/token) 
            & $1.11\times$ (27.94 ms/token) 
            & $1.09\times$ (28.56 ms/token) \\
        Dynamic Policy $\pi_\theta$ 
            & $1.07\times$ (29.02 ms/token) 
            & $1.11\times$ (28.26 ms/token) 
            & $1.07\times$ (29.20 ms/token) \\
        \bottomrule
        \end{tabular}
    }
    \caption{Varying temperature for assisted generation on \textsc{IWSLT2017 En-De}. $\gamma=7$ for all cases.}
    \label{tab:iwstl-temp}
\end{table*}

\begin{table*}[ht]
    \resizebox{\linewidth}{!}{
        \centering
        \begin{tabular}{l|ccc}
        \toprule
        \textsc{Draft Model} & $T=0.5$ & $T=0.9$ & $T=1$ \\
        \hline
        \multicolumn{4}{c}{\cellcolor[gray]{0.8}{\textbf{Speedup}}} \\
        \hline
        Auto-regressive 
            & \multicolumn{3}{c}{\cellcolor[gray]{0.9}{$1.00\times$ (37.06 ms/token)}} \\
        \texttt{T5-Small} 
            & $0.98\times$ (38.21 ms/token) 
            & $0.98\times$ (38.31 ms/token) 
            & $0.97\times$ (38.60 ms/token) \\
        \texttt{T5-Small-XSUM} 
            & $1.18\times$ (31.31 ms/token) 
            & $1.19\times$ (31.23 ms/token) 
            & $1.21\times$ (30.71 ms/token) \\
        \midrule
        Greedy Policy $\pi_\theta$ 
            & $1.16\times$ (31.93 ms/token) 
            & $1.18\times$ (31.43 ms/token) 
            & $1.17\times$ (31.63 ms/token) \\
        Dynamic Policy $\pi_\theta$ 
            & $1.17\times$ (31.50 ms/token) 
            & $1.18\times$ (31.37 ms/token) 
            & $1.16\times$ (31.88 ms/token) \\
        \bottomrule
        \end{tabular}
    }
    \caption{Varying temperature for assisted generation on \textsc{XSum}. $\gamma=7$ for all cases.}
    \label{tab:xsum-temp}
\end{table*}


\end{document}